\documentclass[sigconf]{acmart}

\usepackage{multirow}
\usepackage{threeparttable}
\usepackage{xcolor}
\usepackage{colortbl}
\usepackage{makecell}
\usepackage{booktabs}

\definecolor{softblue}{RGB}{173,216,230}
\usepackage{gradient-text}
\usepackage{xspace}
\newcommand\NickName{{\gradientRGB{Video-CoT}{255, 192, 0}{216, 116, 255}}\xspace}
\definecolor{myorange}{RGB}{255, 192, 0}
\definecolor{mypurple}{RGB}{216, 116, 255}
\AtBeginDocument{%
  }

\setcopyright{acmlicensed}
\copyrightyear{2025}
\acmYear{2025}
\acmDOI{10.1145/nnnnnnn.nnnnnnn}
\acmConference[ACM MM]{Make sure to enter the correct
  conference title from your rights confirmation email}{October 27--31,
  2025}{Dublin, Ireland}
\acmISBN{978-x-xxxx-xxxx-x/YYYY/MM}
\begin{document}

\title{\NickName: A Comprehensive Dataset for Spatiotemporal Understanding of \textcolor{myorange}{Videos} Based on \textcolor{mypurple}{Chain-of-Thought}}


\author{Shuyi Zhang}

\authornote{Both authors contributed equally to this research.}
\affiliation{%
  \institution{Institute of Automation, CAS
School of Artifcial Intelligence, UCAS}
  \city{Beijing}
  \country{China}
}
\email{zhangshuyi2024@ia.ac.cn}

\author{Xiaoshuai Hao}
\authornotemark[1]
\affiliation{%
  \institution{Beijing Academy of Artificial Intelligence (BAAI)}
  \city{Beijing}
  \country{China}
}
\email{xshao @baai.ac.cn}

\author{Yingbo Tang}
\affiliation{%
  \institution{Institute of Automation, CAS
School of Artifcial Intelligence, UCAS}
  \city{Beijing}
  \country{China}
}
\email{tangyingbo2020@ia.ac.cn}

\author{Lingfeng Zhang}
\affiliation{%
  \institution{Shenzhen International GraduateSchool,Tsinghua University}
  \city{Shenzhen}
  \state{Guangdong}
  \country{China}
}
\email{lfzhang715 @ gmail.com}


\author{Pengwei Wang}
\affiliation{%
  \institution{Beijing Academy of Artificial Intelligence (BAAI)}
  \city{Beijing}
  \country{China}
}
\email{pwei@baai.ac.cn}

\author{Zhongyuan Wang}
\affiliation{%
  \institution{Beijing Academy of Artificial Intelligence (BAAI)}
  \city{Beijing}
  \country{China}
}
\email{zyuan@baai.ac.cn}

\author{Hongxuan Ma}
\authornote{Corresponding Authors.}
\affiliation{%
  \institution{Institute of Automation, CAS
School of Artifcial Intelligence, UCAS}
  \city{Beijing}
  \country{China}
}
\email{hongxuan.ma@ia.ac.cn}

\author{Shanghang Zhang}
\authornotemark[2]
\affiliation{%
  \institution{ State Key Laboratory of Multimedia Information Processing, School of Computer Science, Peking University, }
  \city{Beijing}
  \country{China}
}
\email{shanghang@pku.edu.cn}

\renewcommand{\shortauthors}{S.Zhang, X.Hao and Y.Tang et al.}

\begin{abstract}
Video content comprehension is essential for various applications, ranging from video analysis to interactive systems. Despite advancements in large-scale vision-language models (VLMs), these models often struggle to capture the nuanced, spatiotemporal details essential for thorough video analysis.
To address this gap, we introduce \textbf{\textit{\NickName}}, a groundbreaking dataset designed to enhance spatiotemporal understanding using Chain-of-Thought (CoT) methodologies. \NickName contains 192,000 fine-grained spatiotemporal question-answer pairs and 23,000 high-quality CoT-annotated samples, providing a solid foundation for evaluating spatiotemporal understanding in video comprehension.
Additionally, we provide a comprehensive benchmark for assessing these tasks, with each task featuring 750 images and tailored evaluation metrics. Our extensive experiments reveal that current VLMs face significant challenges in achieving satisfactory performance, highlighting the difficulties of effective spatiotemporal understanding.
Overall, the \NickName dataset and benchmark open new avenues for research in multimedia understanding and support future innovations in intelligent systems requiring advanced video analysis capabilities. By making these resources publicly available, we aim to encourage further exploration in this critical area.
Project website: \href{https://video-cot.github.io/}\textbf{\textit{{https://video-cot.github.io/}}}.
\end{abstract}




\maketitle

\begin{figure*}
  \includegraphics[width=0.94\textwidth]{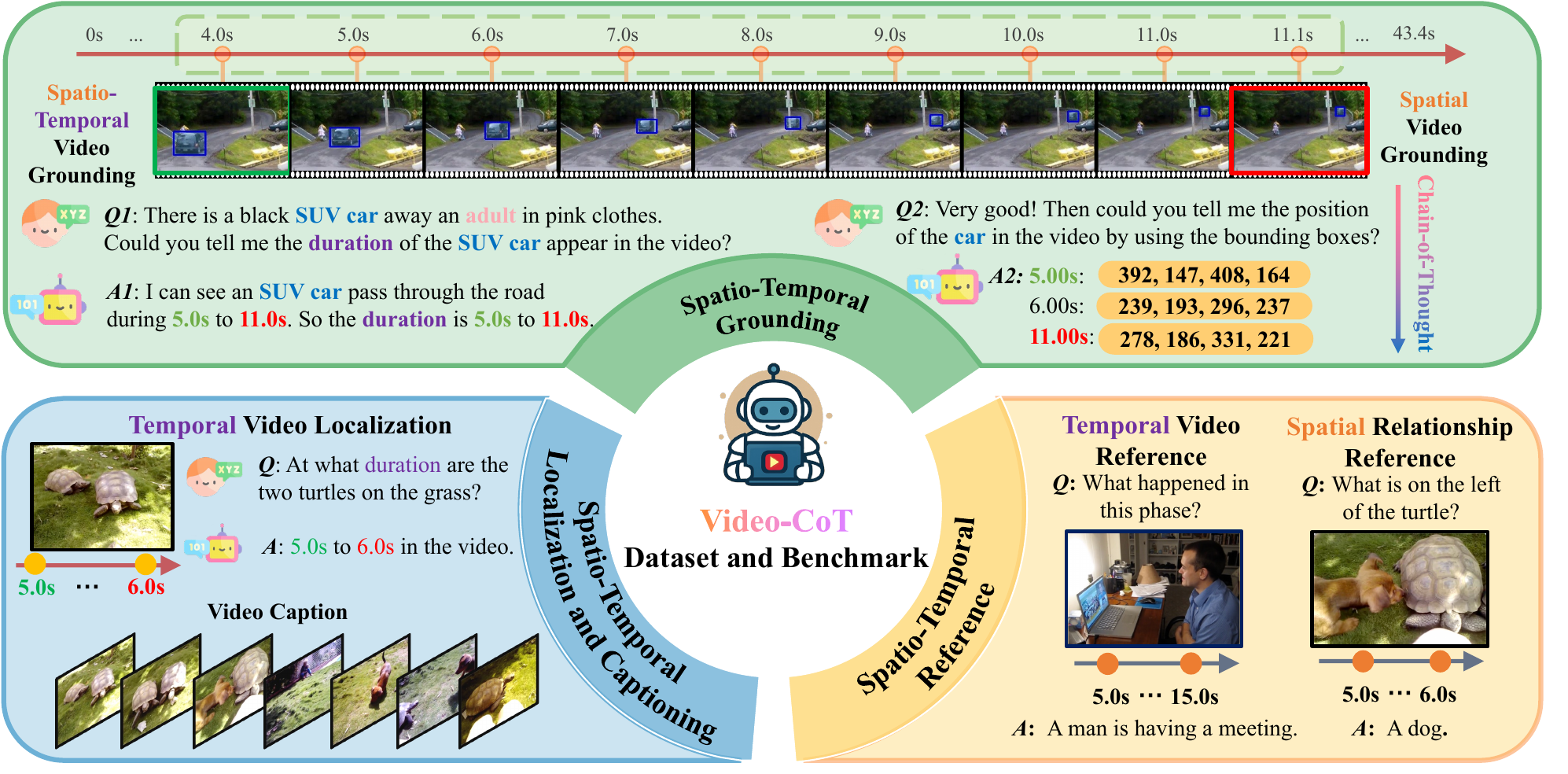}
  \caption{\textbf{Overview of the Video-CoT dataset.} The Video-CoT dataset encompasses three key dimensions: spatio-temporal localization and captioning, spatio-temporal grounding, and spatio-temporal reference.}
  \Description{Enjoying the baseball game from the third-base
  seats. Ichiro Suzuki preparing to bat.}
  \label{fig:figure1}
\end{figure*}

\section{Introduction}
Multimodal content understanding and reasoning are essential for advancing universal artificial intelligence, particularly in the video modality~\cite{hao2023dual, yan2024visa, hao2023uncertainty, shu2023audio, wang2023chatvideo}, where effective comprehension is crucial for applications ranging from automated analysis to interactive systems~\cite{yan2024visa, hao2022listen, hao2021multi,chengshuai}. However, existing video datasets typically focus on content summarization or specific temporal and spatial dimensions, which limits their effectiveness for fine-grained spatiotemporal reasoning. This highlights the need for a task-rich spatiotemporal dataset that facilitates comprehensive video understanding.

Recent advancements in rule-based reinforcement learning have significantly improved the reasoning capabilities of large language models (LLMs), such as GPT-4o~\cite{hurst2024gpt}, Deepseek-R1~\cite{guo2025deepseek}, and Kimi-1.5~\cite{team2025kimi}. This progress has sparked interest in Chain-of-Thought (CoT) data~\cite{wei2022chain}, which incorporates intermediate reasoning steps for problem-solving~\cite{zhang2023multimodal, luo2025ursa}, aiming to enhance complex reasoning abilities by capturing the logical chains of human thought. Although some recent works have generated CoT reasoning data in the video modality~\cite{feng2025video, li2025videochat, zhang2025tinyllava, hao2023mixgen}, they often overlook fine-grained spatiotemporal information—such as event start and end times and the positions of subjects within videos.
Therefore, there is a pressing need to develop a video dataset based on CoT methodologies, addressing these gaps and supporting advanced reasoning in video comprehension.

To address this gap, we introduce \textit{\NickName}, a groundbreaking dataset designed to enhance spatiotemporal understanding using Chain-of-Thought (CoT) methodologies, one example of which is shown in Fig.~\ref{fig:figure1}. \NickName contains 192,000 fine-grained spatiotemporal question-answer pairs and 23,000 high-quality CoT-annotated samples, providing a robust resource for training and evaluating VLMs in video comprehension.
The dataset is organized into three components: \textbf{\textit{Spatio-Temporal Localization and Captioning}}, which includes Temporal Video Localization (TVL) and Video Captioning (VC) tasks for temporal localization and content summarization; 
\textbf{\textit{Spatio-Temporal Grounding}}, consisting of Spatial Video Grounding (SVG) and Spatio-Temporal Video Grounding (STVG), delivering accurate temporal and spatial coordinates based on event descriptions; 
and \textbf{\textit{Spatio-Temporal Reference}}, comprising Spatial Relationship Reference (SRR) and Temporal Video Reference (TVR), addressing relational descriptions of event subjects in time and space.
Through these diverse tasks, we aim to enhance the model's reasoning capibilities to spatiotemporal information. Additionally, we provide a comprehensive benchmark for assessing these tasks, each featuring 750 images and tailored evaluation metrics. 
Extensive experiments reveal that current VLMs face significant challenges in achieving satisfactory performance, underscoring the complexities of effective spatiotemporal understanding.
In summary, our contributions are as follows:

\begin{itemize}

 \item \textbf{Proposal of Video-CoT Dataset.} Introduction of a novel dataset designed to enhance spatiotemporal understanding through Chain-of-Thought (CoT) methodologies, offering a solid foundation for evaluating video comprehension.


 \item \textbf{Comprehensive Benchmark and Evaluation Metrics.} Development of a robust benchmark that includes tailored evaluation metrics for assessing the performance of various tasks within the dataset. This benchmark supports three key dimensions: spatiotemporal localization and captioning, spatiotemporal grounding, and spatiotemporal reference, encompassing six distinct subtasks.

 \item \textbf{Extensive Experiments.} Conducting thorough experiments that highlight the challenges faced by current visual language models (VLMs) in achieving effective spatiotemporal comprehension, offering critical insights that can drive innovation and advancement in future research.
\end{itemize}

\section{Related Work}

\textbf{Video Datasets}
Video understanding is a crucial capability for Multimodal Large Language Models (MLLMs), with accurate video datasets serving as their foundation. Early datasets like UCF-101~\cite{soomro2012ucf101} and Sports-1M~\cite{karpathy2014large} initiated action recognition by emphasizing visual feature learning. As the field matured, more complex datasets emerged, including Kinetics-400/700~\cite{kay2017kinetics}, Something-Something~\cite{goyal2017something}, and ActivityNet-Captions~\cite{krishna2017dense}.
However, existing video datasets often fall short in capturing comprehensive spatiotemporal information. Many focus solely on either temporal or spatial features, lacking effective integration. For example, the Jester dataset~\cite{materzynska2019jester} emphasizes gesture recognition over time but offers minimal spatial annotations for hand and body positions. Similarly, the FineGym dataset~\cite{shao2020finegym} provides temporal annotations for actions in gym videos but lacks detailed spatial context regarding the relationships between athletes and equipment.
These gaps underscore the need for new datasets that effectively combine both spatial and temporal dimensions to enhance video understanding research.

\textbf{Chain-of-Thought (CoT)}
Chain-of-Thought (CoT) prompting~\cite{wei2022chain} has emerged as a pivotal technique for enhancing the complex reasoning capabilities of large language models (LLMs). By explicitly decomposing multi-step problems into intermediate reasoning pathways, CoT significantly improves performance on tasks such as mathematical reasoning~\cite{cobbe2021training} and commonsense QA~\cite{talmor2018commonsenseqa}. Subsequent advances focus on three directions: (1) Minimizing demonstration dependency, as seen in Zero-Shot CoT activated by trigger phrases like "Let's think step by step"~\cite{kojima2022large}; (2) Optimizing reasoning robustness through methods like self-consistency (aggregating multiple reasoning paths)~\cite{wang2022self} and automated demonstration generation~\cite{zhang2022automatic}; (3) Integrating external tools such as program interpreters~\cite{gao2023pal} and symbolic solvers to address computational limitations. Despite these innovations, critical challenges persist in ensuring reasoning fidelity—generated chains often contain logical inconsistencies~\cite{hsieh2023distilling} and factual hallucinations~\cite{lyu2023faithful}, motivating ongoing research into verifiable reasoning frameworks and error-correction mechanisms.

\begin{figure}
  \includegraphics[width=0.5\textwidth]{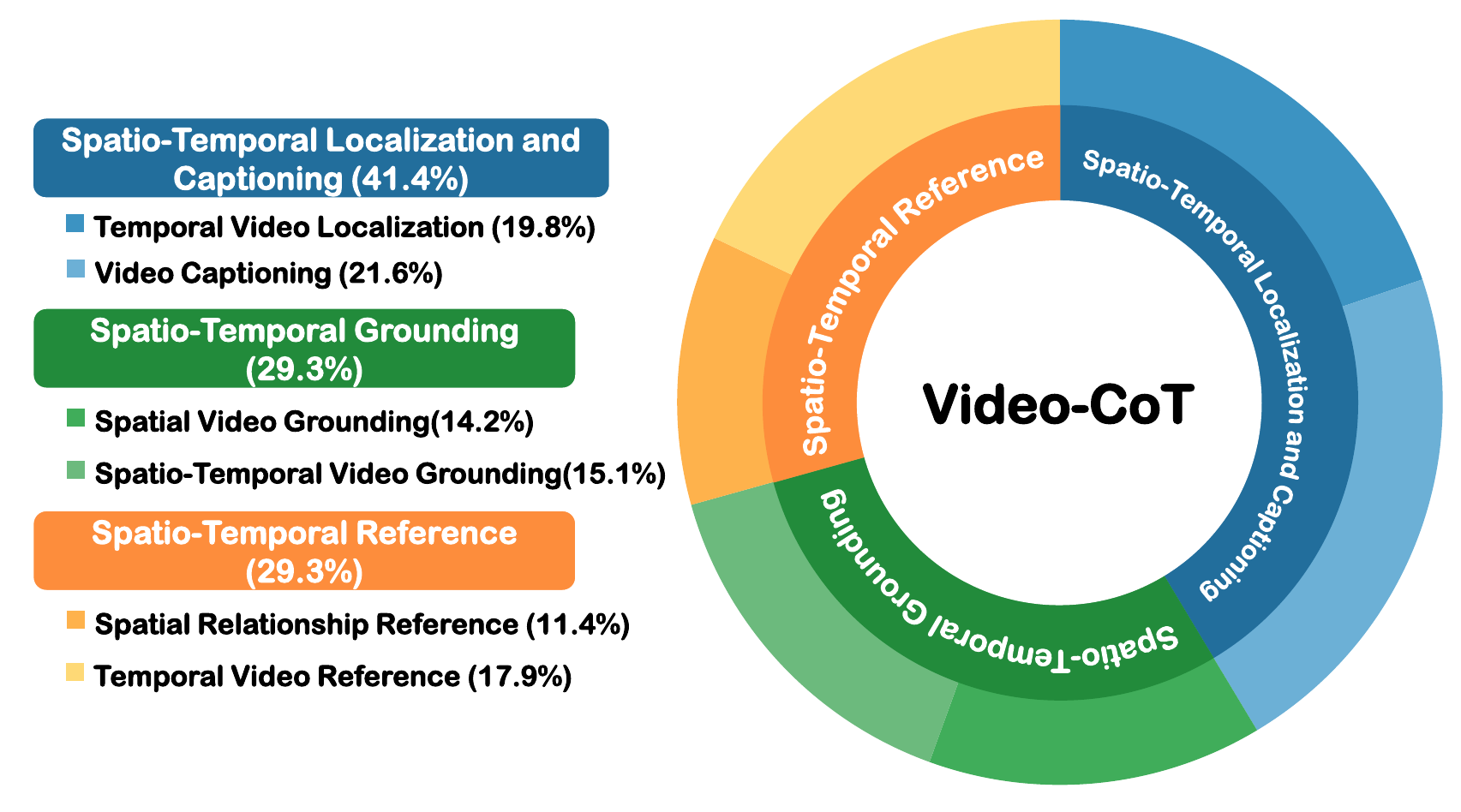}
  \caption{The distribution of tasks across three primary categories.}
  \Description{The distribution of tasks across three categories.}
  \label{fig:data}
\end{figure}

\section{Dataset Construction and Analysis}
In this section, we provide an in-depth exploration of each spatiotemporal task within our dataset and outline the Chain-of-Thought (CoT) data generation process in detail.

\subsection{Task Definition}
Our Video-CoT dataset consists of six tasks  categorized into three key aspects:

\textbf{Spatio-Temporal Localization and Captioning} connects visual perception with language understanding by detecting specific temporal events in videos and generating meaningful textual descriptions. This task establishes the foundation for identifying and articulating key events within video content. It comprises two subtasks: \emph{\textbf{Temporal Video Localization (TVL)}}, which focuses on pinpointing the precise temporal boundaries of actions or events in untrimmed videos, such as identifying the exact duration of a ``basketball dunk'' in a sports clip; and \emph{\textbf{Video Captioning (VC)}}, which extends beyond event detection to generate natural language descriptions of video content, enabling more intuitive interaction with video data across various contexts.


\begin{figure}
  \includegraphics[width=0.48\textwidth]{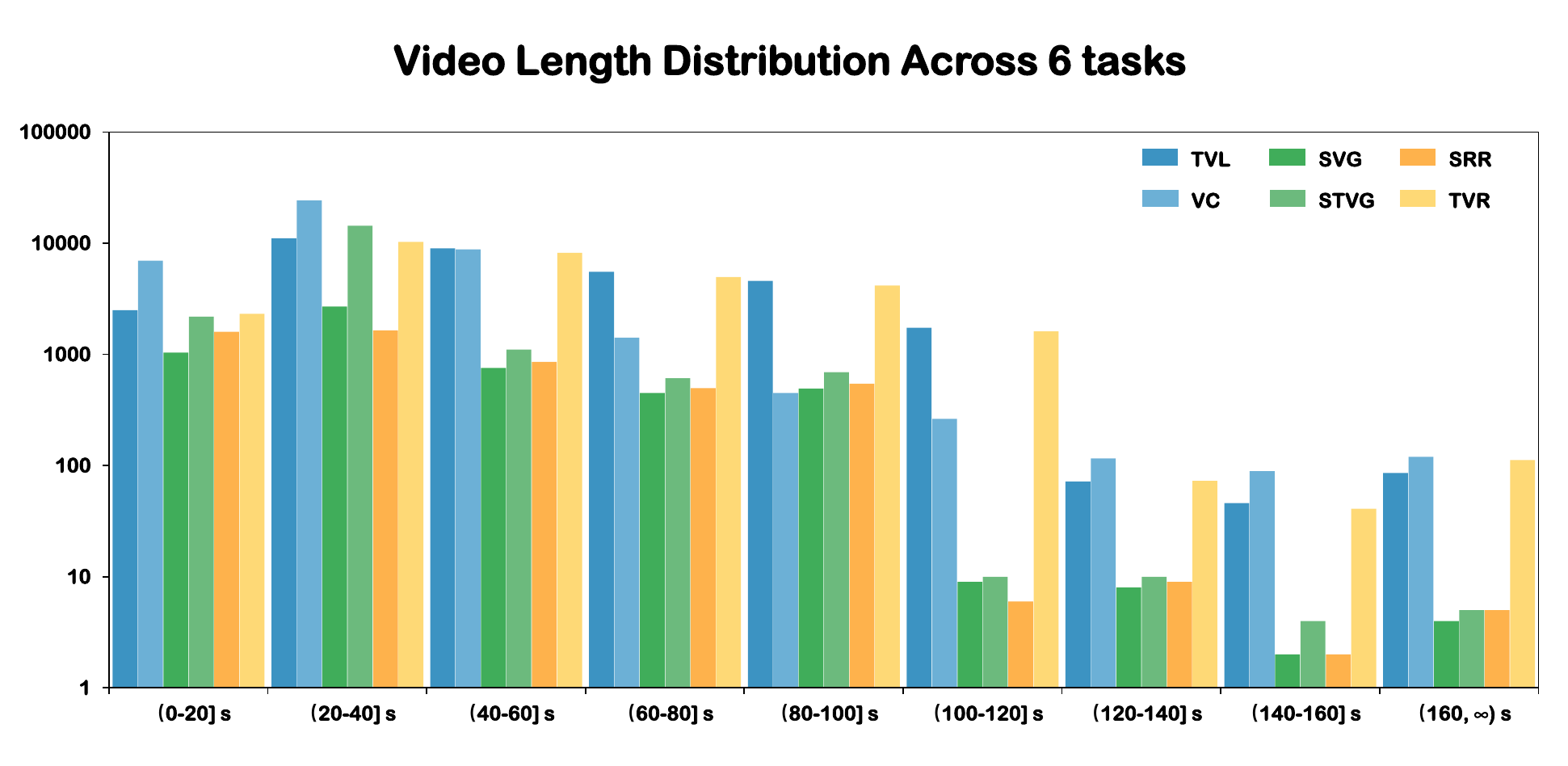}
  \caption{The video length statistic.}
  \Description{The video length statistic.}
  \label{fig:time}
\end{figure}

\begin{figure*}
\centering
  \includegraphics[width=\textwidth]{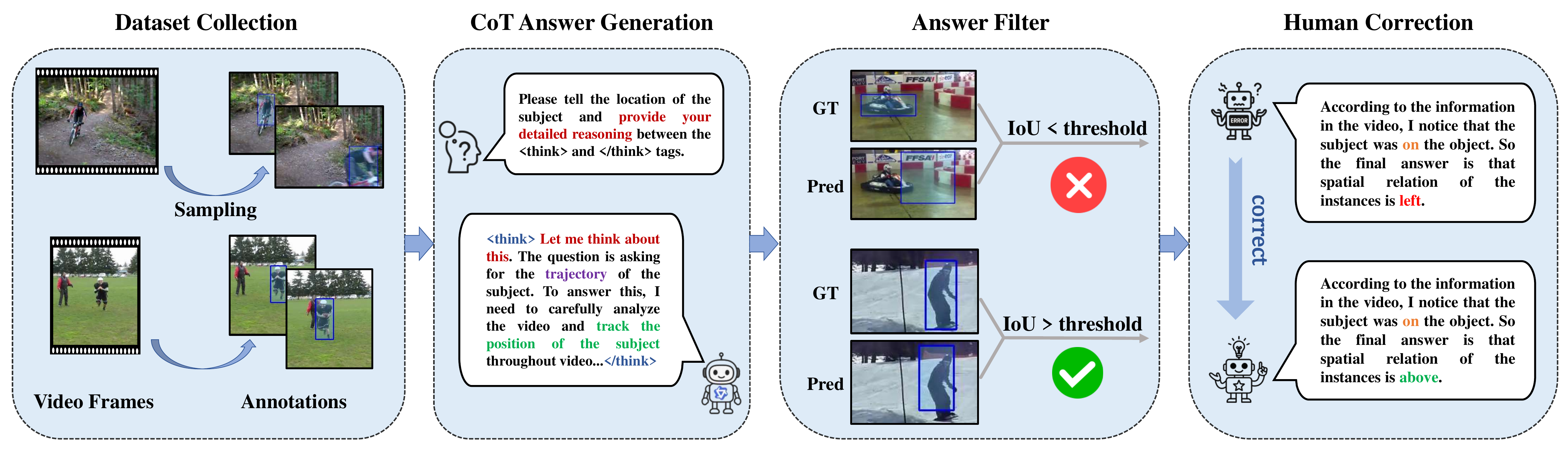}
  \caption{Pipeline for Constructing the Video-CoT Dataset.}
  \label{fig:pipeline}
\end{figure*}

\textbf{Spatio-Temporal Grounding} aims to align textual queries with corresponding spatio-temporal regions in videos, integrating spatial and temporal reasoning to enhance multimodal understanding. This task includes two components: \emph{\textbf{Spatial Video Grounding (SVG)}}, which focuses on locating specific spatial regions in video frames that correspond to a textual query, such as identifying a ``black SUV car leaving an adult in pink clothes''; and \emph{\textbf{Spatio-Temporal Video Grounding (STVG)}}, which extends SVG by predicting a spatio-temporal sequence of bounding boxes across multiple frames that match a text query, like tracking ``a dog chasing a ball''. STVG integrates spatial localization with temporal tracking, addressing challenges like object occlusion and appearance variations to ensure robust alignment between textual and visual representations.


\textbf{Spatio-Temporal Reference} focuses on understanding object relationships and temporal dynamics within video content. By modeling spatial and temporal dependencies, this task enhances the interpretation of complex interactions in dynamic scenes. It comprises two key components: \emph{\textbf{Spatial Relationship Reference (SRR)}}, which identifies spatial relationships between objects, such as ``above'', ``behind'', or ``next to'' within video scenes— for example, detecting ``a book on a table'' during a room scan, crucial for precise scene understanding; and \emph{\textbf{Temporal Video Reference (TVR)}}, which describes video content within specific time intervals. Given a defined segment, TVR generates concise and contextually accurate textual descriptions of events and actions occurring during that interval, requiring a deep understanding of temporal dependencies and event dynamics to produce temporally coherent and meaningful descriptions.


\subsection{CoT Dataset Building}

\textbf{Data Collection and Curation} for our Video-CoT dataset primarily relies on publicly available datasets. For the tasks of Temporal Video Localization (TVL), Video Captioning (VC), and Temporal Video Reference (TVR), we directly utilize existing datasets (VTimeLLM-stage2~\cite{huang2024vtimellm}, MSR-VTT~\cite{xu2016msr}, MSVD~\cite{chen2011collecting} and WebVid~\cite{bain2021frozen}). In contrast, the tasks of Spatial Relationship Reference (SRR), Spatial Video Grounding (SVG), and Spatio-Temporal Video Grounding (STVG) involve additional processing based on the HCSTVG-V1~\cite{tang2021human} and VidSTG~\cite{zhang2020does} datasets, which provide dense annotations for each frame. 

Our filtering pipeline selects subjects or objects as query targets and ensures their consecutive presence across video frames, discarding objects that appear for less than 2 seconds. We derive the start and end timestamps of objects with continuous presence, along with their pixel locations at initial, final, and intermediate integer-second timestamps based on the video's frame counts and per-frame annotations. By maintaining the spatial relationships between primary event subjects and other objects in the environment, we construct spatiotemporal relational question-answering data, resulting in a total of 192K fine-grained spatial-temporal QA pairs.

The distribution of our dataset across six tasks is illustrated in Fig.~\ref{fig:data}. Additionally, the dataset demonstrates a balanced distribution across various video length intervals, as shown in Fig.~\ref{fig:time}. Short videos (0–40 seconds) comprise the majority, providing ample data for tasks that require rapid event detection and description. Simultaneously, the dataset includes a significant number of longer videos exceeding 160 seconds, enabling models to train on complex long-term temporal sequences while maintaining performance. This distribution reflects real-world video content characteristics, encompassing both short-term events and extended sequences, thereby supporting the development of robust models capable of generalizing across diverse video lengths and inference tasks.

\textbf{CoT Data Generation} We construct a pipeline for Chain-of-Thought (CoT) data generation in the Video-CoT dataset, as illustrated in Fig.~\ref{fig:pipeline}. Leveraging existing video datasets and annotations, we employ a two-stage methodology utilizing a vision-language model (VLM) to analyze spatiotemporal relationships. Specifically, we utilize the Qwen2.5-VL-72B-Instruct model, known for its exceptional ability to generate coherent, long-form reasoning chains. The VLM processes input prompts alongside corresponding video data while referencing relevant examples, producing structured reasoning outputs encapsulated within \texttt{<think>} and \texttt{</think>} tags to represent the complete thought process.

To ensure the quality of the generated CoT data, we apply task-specific evaluation metrics for assessing the answers. We establish thresholds based on the average accuracy of individual tasks and the best-performing results from prior studies, ensuring that the model's outputs surpass the average performance of state-of-the-art models. Only samples with answers exceeding these thresholds are retained in the Video-CoT dataset, effectively filtering out low-quality chains of thought that may contain logical inconsistencies or factual errors.
To further enhance dataset quality, we implement a manual expert review process to eliminate any remaining low-quality data. This two-stage quality control mechanism—combining automated screening based on metrics and manual expert verification—ensures that our Video-CoT dataset comprises only high-quality chain-of-thought samples, thereby effectively improving the model's spatiotemporal reasoning capabilities.


\begin{table*}[t]
  \centering
  \caption{Performance Comparison of VLMs on Video-CoT Benchmark Tasks.}
  \fontsize{4.5}{5.5}\selectfont 
  \resizebox{0.9\linewidth}{!}{
      \begin{threeparttable}
          \begin{tabular}{l|c|ccccc|cc}
              \toprule
              \multirow{2}{*}{\textbf{Models}} & \multirow{2}{*}{\textbf{Params.}} & \textbf{TVL} & \textbf{VC} & \textbf{SRR} & \textbf{TVR} & \textbf{SVG} & \multicolumn{2}{c}{\textbf{STVG}}\\
              \cmidrule{3-9}
              & & tIoU$\uparrow$ & MENTOR$\uparrow$  & EM$\uparrow$ & MENTOR$\uparrow$  & sIoU$\uparrow$ & tIoU$\uparrow$ & sIoU$\uparrow$ \\                                     
              \midrule
              \multicolumn{9}{c}{\textit{Closed-Source MLLMs}} \\
              \midrule
              \rowcolor{red!8}GPT-4o~\cite{hurst2024gpt}& $-$ & 21.7&26.9 &14.6 &25.0 & 15.9& 14.3 & 7.4 \\
              \rowcolor{red!8}Gemini-1.5-pro~\cite{team2024gemini}       & $-$& 24.2& 29.0& 12.8& 27.3& 20.9& 15.8& 9.9\\
              \midrule
              \multicolumn{9}{c}{\textit{Open-Source MLLMs}} \\
              \midrule
              \rowcolor{yellow!8}Qwen2.5-VL~\cite{bai2025qwen2}          & 3B & 4.4 & 4.6 & 10.0 & 14.8& 10.3 & 8.5 & 5.7 \\
              \rowcolor{yellow!8}Qwen2.5-VL~\cite{bai2025qwen2}          & 7B & 20.0 & 19.1 & 12.4 & 20.1& 12.5 & 12.9 & 5.9 \\
              \rowcolor{yellow!8}InternVideo2\_5\_Chat~\cite{wang2025internvideo2}         & 8B & 11.8 & 20.1 & 7.1 & 3.4 & 9.4 & 10.4 & 5.4 \\
              \rowcolor{yellow!8}Video-UTR~\cite{yu2025unhackable}            & 7B & 10.3 & 3.7 & $-$ & $-$ & 2.9 & $-$ & $-$\\
              \rowcolor{yellow!8}LLaVA-NeXT-Video~\cite{zhang2024llava}            & 7B & 8.5 & 4.3 & 6.6 & $-$ & $-$ & $-$ & $-$\\
              \rowcolor{yellow!8}LLaVA-onevision-qwen2~\cite{li2024llava}       & 7B & 10.4 & 14.2& 9.5 & 12.8 & 4.5 & 13.0 & 4.9\\
              \bottomrule
          \end{tabular}
      \end{threeparttable}
  }
  \label{tab:main_results}
  \vspace{-0.5em}
\end{table*}

\begin{table*}[!t]
  \centering
  \caption{Performance of Our Proposed Models on Video-CoT Benchmark Tasks.}
  \fontsize{4.5}{5.5}\selectfont 
  \resizebox{0.9\linewidth}{!}{
      \begin{threeparttable}
          \begin{tabular}{l|c|ccccc|cc}
              \toprule
              \multirow{2}{*}{\textbf{Models}} & \multirow{2}{*}{\textbf{Params.}} & \textbf{TVL} & \textbf{VC} & \textbf{SRR} & \textbf{TVR} & \textbf{SVG} & \multicolumn{2}{c}{\textbf{STVG}}\\
              \cmidrule{3-9}
              & & tIoU$\uparrow$ & MENTOR$\uparrow$  & EM$\uparrow$ & MENTOR$\uparrow$  & sIoU$\uparrow$ & tIoU$\uparrow$ & sIoU$\uparrow$ \\                                     
              \midrule
              Qwen2.5-VL~\cite{bai2025qwen2} (Baseline)          & 3B & 4.4 & 4.6 & 10.0 & 14.8& 10.3 & 8.5 & 5.7 \\
              \cellcolor{blue!6}\textbf{Video-Ans-SFT (Ours)} & \cellcolor{blue!6}3B & \cellcolor{blue!6}{8.3 (+3.9$\uparrow$)}& \cellcolor{blue!6}{12.3 (+7.7$\uparrow$)}& \cellcolor{blue!6}{14.1 (+4.1$\uparrow$)}& \cellcolor{blue!6}\textbf{22.8 (+8.0$\uparrow$)}& \cellcolor{blue!6}{14.2 (+3.9$\uparrow$)}& \cellcolor{blue!6}{12.2 (+3.7$\uparrow$)}& \cellcolor{blue!6}{8.1 (+2.4$\uparrow$)}\\
             \cellcolor{blue!10}\textbf{Video-CoT-SFT (Ours)} & \cellcolor{blue!10}3B & \cellcolor{blue!10}\textbf{19.7 (+15.3$\uparrow$)}& \cellcolor{blue!10}\textbf{15.4 (+10.8$\uparrow$)}& \cellcolor{blue!10}\textbf{16.9 (+6.9$\uparrow$)}& \cellcolor{blue!10}{21.1 (+6.3$\uparrow$)}& \cellcolor{blue!10}\textbf{17.0 (+6.7$\uparrow$)}& \cellcolor{blue!10}\textbf{14.1 (5.6$\uparrow$)}& \cellcolor{blue!10}\textbf{9.2 (3.5$\uparrow$)}\\
                  
              \bottomrule
          \end{tabular}
      \end{threeparttable}
  }
  \label{tab:our_models_results}  
\end{table*}

\section{Method}

To enhance the spatiotemporal reasoning capabilities of the vision-language model (VLM) and validate the effectiveness of the \textbf{\textit{\NickName dataset}}, we propose a comprehensive fine-tuning framework comprising two methods: Answer Supervised Fine-tuning (Ans-SFT) and Chain of Thoughts Supervised Fine-tuning (CoT-SFT). These methods utilize distinct data types from the original dataset and the \textbf{\textit{\NickName dataset}} to fine-tune the same model, optimizing performance across various levels of spatiotemporal understanding. In the following sections, we provide detailed descriptions of these two fine-tuning methods, including their model architectures, input-output configurations, and loss function formulations.

\subsection{Answer-Supervised Fine-Tuning (Ans-SFT)}
\label{subsec:ans-sft}
Ans-SFT focuses on direct answer alignment through end-to-end optimization of answer generation probability. The input-output formulation is expressed as:
\begin{equation}
\text{Input: } {V, Q} \rightarrow \text{Output: } A,
\end{equation}
where $V \in \mathbb{R}^{T \times H \times W \times C}$ represents the input video with $T$ frames, and $Q$ denotes the natural language question. Given this input pair, the model $\mathcal{M}$ generates the target answer $A$ by maximizing the conditional likelihood $P(A \mid V,Q$):
\begin{equation}
A^* = \arg\max_A P(A \mid V, Q; \theta).
\end{equation}
The method employs standard cross-entropy loss for sequence modeling:
\begin{equation}
\mathcal{L}_{\text{Ans}} = -\sum_{t=1}^{|A|} \log P(a_t \mid a_{<t}, V, Q).
\end{equation}
This approach is effective for spatiotemporal localization tasks where answers correspond to specific spatiotemporal coordinates (\textit{e.g.}, timestamps or bounding boxes). Direct SFT enables rapid convergence and efficient training for tasks requiring precise localization. However, it may face limitations in handling tasks that demand fine-grained analysis or multi-step reasoning, such as Spatio-Temporal Video Grounding (STVG) or Spatial Relationship Reference (SRR), where intermediate reasoning steps are crucial for achieving accurate results.


\subsection{Chain-of-Thought-Supervised Fine-Tuning (CoT-SFT)}
\label{subsec:cot-sft}
CoT-SFT introduces hierarchical reasoning supervision by leveraging structured reasoning chains from our \textbf{\textit{\NickName dataset}}. The input-output formulation expands the basic framework to include intermediate reasoning steps:
\begin{equation}
\text{Input: } {V, Q} \rightarrow \text{Output: } {R, A},
\end{equation}
where \( R = \{r_1, r_2, \ldots, r_n\} \) represents the annotated reasoning chain with \( n \) intermediate steps, and \( A \) denotes the final answer. The model learns to generate both the reasoning process and the ultimate solution through a two-phase optimization objective:
\begin{equation}
\begin{aligned}
\mathcal{L}_{\text{CoT}} &= -\sum_{t=1}^{|R|} \log P(r_t \mid r_{<t}, V, Q; \theta) \\
&\quad - \lambda \sum_{t=1}^{|A|} \log P(a_t \mid a_{<t}, R, V, Q; \theta),
\end{aligned}
\end{equation}

where \( \lambda \) is a hyperparameter that balances the importance of reasoning step generation and final answer prediction. This dual-objective formulation enables the model to simultaneously learn explicit reasoning patterns and accurate answer generation.

This approach is particularly effective for complex multi-step tasks where intermediate reasoning processes—such as object trajectory analysis, event sequencing, or spatial relationship inference—are crucial for achieving correct solutions. To enhance training efficiency and model robustness, we implement a curriculum learning strategy that progressively increases reasoning chain complexity. The curriculum begins with simple 2-step reasoning scenarios (\textit{e.g.}, temporal localization followed by spatial identification) and advances to sophisticated 5-step chains involving complex spatial relationship deduction and causal inference.

\section{Experiments}

\subsection{Environment Setup}

\textbf{Benchmark}  
To evaluate the effectiveness of our proposed methods and dataset, we constructed the Video-CoT-Benchmark using the same methodology as the Video-CoT dataset. This comprehensive benchmark encompasses all six spatiotemporal understanding sub-tasks: Video Captioning (VC), Temporal Video Localization (TVL), Spatial Video Grounding (SVG), Spatio-Temporal Video Grounding (STVG), Spatial Relationship Reference (SRR), and Temporal Video Reference (TVR). The benchmark includes 4,500 video question-answer pairs that are entirely distinct from the training dataset, ensuring a thorough and unbiased evaluation of model performance across various spatiotemporal reasoning tasks.


\textbf{Implementation Details}  
We conduct our experiments using the Qwen2.5-VL-3B-Instruct model as the baseline, fine-tuning it through two distinct methodologies: Answer-Supervised Fine-Tuning (Ans-SFT) and Chain-of-Thought-Supervised Fine-Tuning (CoT-SFT). The fine-tuning process utilizes the AdamW optimizer with a learning rate of \(10^{-6}\), employing mixed-precision training to enhance efficiency on a single GPU. Training is performed for 1 epoch to ensure effective learning from the dataset while minimizing the risk of overfitting.


\textbf{Evaluation Metrics}  
We established distinct evaluation metrics for each task. For Temporal Video Localization (TVL), we used temporal Intersection over Union (tIoU) to assess localization accuracy. In Video Captioning (VC) and Temporal Video Reference (TVR), we employed the MENTOR metric to evaluate the quality and relevance of generated captions and textual descriptions. For Spatial Video Grounding (SVG) and Spatio-Temporal Video Grounding (STVG), we utilized spatial Intersection over Union (sIoU) to measure precision in spatial localization within single frames and across frames, respectively, with STVG also incorporating tIoU to ensure temporal accuracy. For Spatial Relationship Reference (SRR), Exact Match (EM) was used to verify if predicted spatial relationships align with ground truth labels. These criteria were selected based on their relevance to each task's requirements and their effectiveness in measuring model performance in capturing spatiotemporal information.


\subsection{Results}
The evaluation on Video-CoT benchmark reveals significant limitations in current state-of-the-art VLMs when performing fine-grained spatiotemporal reasoning for videos, as shown in Table~\ref{tab:main_results}. Closed-source models generally outperform open-source counterparts across all benchmark tasks, with Gemini-1.5-pro achieving the highest scores in most metrics. However, even the advanced models demonstrate limited performance, particularly in STVG task where the best score of 9.9 sIoU remains relatively low.

Table ~\ref{tab:our_models_results} demonstrates the effectiveness of our proposed training approaches compared to the baseline Qwen2.5-VL (3B) model. Both the Video-Ans-SFT and Video-CoT-SFT methods yield substantial improvements across all tasks, with the CoT-SFT variant showing particularly strong gains in temporally-oriented metrics (tIoU) for TVL and TVR tasks. The Video-CoT-SFT achieves a remarkable 19.7 tIoU in TVL, outperforming not only its 3B baseline but also all other open-source 7B models listed in Table~\ref{tab:main_results}.  This performance advantage confirms that Chain-of-Thought fine-tuning specifically enhances a model's capacity for long-term spatiotemporal reasoning compared to standard answer-based supervision, providing a promising direction for developing more capable models.

\section{Potential Applications}  
The Video-CoT dataset, with its dense spatiotemporal annotations, balanced distribution of short and long videos, and structured Chain-of-Thought (CoT) annotations, enables advanced research across multiple domains. We highlight four key application areas that leverage its unique characteristics:

\textbf{Training and Benchmarking Spatiotemporal Reasoning}  
The Video-CoT dataset offers dense spatiotemporal annotations and object trajectory information, providing high-quality benchmark data for video spatiotemporal relation reasoning tasks~\cite{wang2025internvideo2,wuevaluating,yu2025unhackable,zhang2024llava}. It supports the training of vision-language models (VLMs) with spatiotemporal reasoning capabilities, including understanding relationships and event progression.
The structured Chain-of-Thought annotations (\texttt{<think>}...\texttt{</think>}) enhance the development of interpretable video reasoning models, applicable in areas such as autonomous driving (analyzing vehicle-pedestrian interactions) and surveillance systems (detecting anomalous spatiotemporal patterns). Additionally, its per-frame object localization and duration labels enable precise evaluation of relational reasoning accuracy.

\textbf{Advancing Long-term Video Understanding}  
The dataset features videos exceeding 160 seconds in duration, allowing researchers to develop innovative long-term temporal modeling approaches~\cite{ weng2024longvlm,zhang2025mapnav,zhang2024multi}. These extended sequences support research on long-range dependency modeling and event detection, particularly in applications requiring state tracking, such as sports commentary.
Moreover, the balanced mix of short and long videos enhances studies on temporal generalization.



\textbf{Cross-modal Retrieval Enhancement}  
The comprehensive annotations in Video-CoT significantly advance video-text retrieval systems~\cite{hao2023dual,hao2021matters,hao2022listen,hao2021multi,hao2023uncertainty,zhusavideo} by enabling cross-modal search capabilities. The dataset supports fine-grained queries that incorporate spatial, temporal, and relational constraints, such as identifying specific object interactions within defined time windows. This level of detail facilitates accurate temporal grounding of textual descriptions to specific video segments and enhances compositional cross-modal retrieval through advanced relational reasoning. These capabilities hold transformative potential in forensic investigations, allowing analysts to efficiently search through extensive video evidence.


\textbf{Intelligent Human-Computer Interaction Systems}
The precise spatiotemporal annotations in Video-CoT create new opportunities for intelligent interactive systems~\cite{pang2025hierarchical,tang2025affordgrasp} requiring nuanced video understanding. In augmented and virtual reality applications, the dataset's detailed object trajectory data facilitates more natural interactions between virtual and real-world objects, ensuring proper occlusion effects. 
For robotic systems, timestamped object positions and interaction patterns enhance dynamic environment mapping, enabling robots to anticipate human actions effectively. The structured \texttt{<think>} annotations are especially valuable for developing explainable AI assistants, allowing them to articulate their spatiotemporal reasoning processes during human-computer interactions and bridging the gap between complex visual understanding and intuitive communication.

\section{Conclusion}
This paper introduces \textit{\NickName}, a novel dataset aimed at enhancing spatiotemporal understanding in video comprehension using Chain-of-Thought (CoT) methodologies. It contains 192,000 fine-grained spatiotemporal question-answer pairs and 23,000 high-quality CoT-annotated samples, organized into three key components. A comprehensive benchmark for evaluation is also provided. Our experiments reveal significant challenges faced by current vision-language models (VLMs). In summary, \NickName represents a valuable contribution to the community, advancing research in multimodal content understanding.

\clearpage
\bibliographystyle{ACM-Reference-Format}
\bibliography{sample-base}










\end{document}